\documentclass[a4paper,fleqn]{cas-dc}

\usepackage[authoryear]{natbib}
\usepackage{algpseudocode}
\usepackage{algorithm}
\usepackage{multirow}

\def\tsc#1{\csdef{#1}{\textsc{\lowercase{#1}}\xspace}}
\tsc{WGM}
\tsc{QE}
\tsc{EP}
\tsc{PMS}
\tsc{BEC}
\tsc{DE}

\begin{document}
\let\WriteBookmarks\relax
\def\floatpagepagefraction{1}
\def\textpagefraction{.001}

\shorttitle{Uncertainty of  canopy height  Estimation from GEDI LiDAR}

\shortauthors{Jose Bermudez et~al.}

\title [mode = title]{A Deep Learning Approach to Estimate Canopy Height and Uncertainty by Integrating Seasonal Optical, SAR and Limited GEDI LiDAR Data over Northern Forests}
 
\tnotemark[1,2]




%







\affiliation[1]{organization={McMaster University},
    addressline={
    1280 Main Street West}, 
    city={Hamilton},
    postcode={L8S 4K1}, 
    state={Ontario},
    country={Canada}}

\affiliation[2]{organization={Toronto Metropolitan University},
    addressline={350 Victoria Street}, 
    city={Toronto},
    postcode={M5B 2K3}, 
    state={Ontario},
    country={Canada}}

\affiliation[3]{organization={Planet Labs PBC},
    city={San Francisco},
    postcode={695571}, 
    state={Californa},
    country={USA}}


\affiliation[4]{Environment and Climate Change Canada}

\author[1]{Jose Bermudez}[type=editor,
                        auid=000,
                        bioid=1,
                        ]


\cormark[1]


\ead{bermudej@mcmaster.ca}

\author[1,2]{Cheryl Rogers}
\ead{cheryl.a.rogers@torontomu.ca}

\author[3]{Camile Sothe}[%
   ]
\ead{camile.sothe@planet.com }


\author%
[4]
{Dominic Cyr}
\ead{dominic.cyr@ec.gc.ca}

\author%
[1]
{Alemu Gonsamo}
\ead{gonsamoa@mcmaster.ca}

\cortext[cor1]{Corresponding author}
\cortext[cor2]{Principal corresponding author}



\begin{abstract}
Accurate estimation of forest canopy height is crucial for assessing aboveground biomass and carbon stock dynamics to support ecosystem monitoring and associated services such as timber provisioning, climate change mitigation and biodiversity conservation efforts. Despite significant advancements in spaceborne LiDAR observations in recent years, data for northern high latitudes remains scarce due to orbital and sampling limitations. This study presents a methodology for developing spatially continuous high-resolution canopy height and uncertainty estimates using Deep Learning Regression models. Our approach leverages multi-source and multi-seasonal satellite observations from Sentinel-1, Landsat, and ALOS-PALSAR-2 sensors, and uses spaceborne GEDI LiDAR observations as reference. Experimental results conducted in Ontario, Canada, and evaluated against airborne LiDAR data, demonstrate the effectiveness of the proposed approach. Our models achieved the best performance, that is comparable to GEDI data alone, when seasonal features from Sentinel-1 and Landsat bands, combined with PALSAR data are used in model training. Specifically, the best-performing model achieved an \textsuperscript{2} of 0.72, with an RMSE of 3.43m and a bias of 2.44m. Using seasonal data instead of summer-only data increased explained variability by 10\%, reduced canopy height error by 0.45m, and decreased bias by 1m. The weighting strategy applied in the deep learning model proved more effective in reducing the error of tall canopy height estimation compared to a recent global model, although this led to overestimation for low canopy heights. The uncertainty map indicated higher uncertainty for estimates close to forest edge where GEDI observations can present measurement errors and SAR backscatter foreshortening, layover and shadow are expected. This study advances canopy height estimation methodologies for regions not covered by spaceborne LiDAR sensors, providing valuable tools for various remote sensing applications in forestry, environmental monitoring, and carbon stock estimation. 


\end{abstract}


\begin{highlights}

\item We synthesized a deep learning model and weighting scheme for canopy height and uncertainty estimation.
\item Canopy height is estimated from GEDI, seasonal optical and SAR data.
\item Canopy height is estimated for high latitude forests where GEDI data is scarce.
\item The use of seasonal data decreased the estimated canopy height error by 0.45m .
\item Highest uncertainty of canopy height estimation was found close to forest edge. 

\end{highlights}

\begin{keywords}
LIDAR \sep PALSAR-2 \sep Landsat \sep Sentinel-1 \sep Canopy Height \sep ResUNet \sep  Uncertainty \sep Boreal Forests
\end{keywords}

\maketitle

\section{Introduction}

Accurate estimates of forest aboveground biomass (AGB) are essential for assessing forest carbon stocks and their change over time to support policies for climate change mitigation, resource management, and biodiversity conservation (\cite{asner2014functional,mitchard2014markedly, stovall2021comprehensive}). Canopy height, a variable that can be directly obtained from remote sensing observations, is used as a predictor of AGB in many studies because of their close relationship (\cite{asner2014functional, jucker2017allometric, carreiras2017coverage, csillik2019monitoring, qi2019forest, hentz2018estimating, silva2021treetop}. Canopy height is also critical to accurately estimate carbon removal rates, as the rate of carbon assimilation into biomass increases continuously with the size of trees (\cite{stephenson2014rate}). 

Airborne Light Detection and Ranging (LiDAR) is the preferred means to provide highly accurate and precise measurements of canopy height and unlike field observations, allows for height measurements to be taken from remote and larger areas in a timely manner. However, airborne LiDAR (ALS) campaigns have limited spatial coverage and tend to be expensive, leading most of the acquisitions to focus on areas of high-value forests (\cite{wulder2012lidar, tompalski2021estimating}). In this sense, the Global Ecosystem Dynamics Investigation (GEDI) spaceborne LiDAR sensor onboard the International Space Station (ISS) offers a unique opportunity to overcome this challenge. The GEDI mission uses a laser altimeter to measure the distance from the ISS to the Earth’s surface with high accuracy and spatial resolution (\cite{dubayah2020global}). GEDI does not provide a spatially continuous canopy height map. Instead, it captures 25 m spatial resolution footprint samples over the Earth’s surface following a sparse-grid-based sampling pattern between 51.6° N and 51.6° S. GEDI samples are spaced every 60 m in the along-track direction and 600 m in the across-track direction (\cite{liu2021performance}). 

To extrapolate GEDI LiDAR measurements and build wall-to-wall canopy height maps, many studies make use of other sensors data and machine learning methods. For example, \cite{qi2019forest} used GEDI to improve height estimates provided by TanDEM-X InSAR in sites located in the United States and Costa Rica. At a national scale, \cite{sothe2022spatially} used both ICESat-2 and GEDI LiDAR observations together with Sentinel-1 and -2, PALSAR-2 data, and a random forest model to map forest canopy height in Canada at 250 m spatial resolution with mean error of 4.2 m. At a global scale, \cite{potapov2021mapping} used GEDI canopy height data with Landsat data and a tree ensemble model to produce a 30 m spatial resolution forest canopy height map with mean error of 9.1 m. More recently,\cite{travers2024mapping} used satellite LiDAR from ICESat-2, Landsat data and a random forest model to map canopy height of northern forest-tundra areas in Canada. 

A common limitation with tree ensemble methods like random forest is that predictions are resulting from the average values of many decision trees, which undermines their ability to predict canopy height for very short or tall canopy height values \cite{wright2017unbiased, travers2024mapping}. Most importantly, tree ensemble methods exhibit high sensitivity of predictions to input data quality, poor performance with a small training dataset, and failure to extrapolate prediction beyond the bound of the training dataset \cite{hengl2018random}. 

Recent studies indicate that deep learning (DL) methods can enhance the accuracy of canopy height maps compared to conventional machine learning approaches due to their ability to learn complex patterns without relying on hand-crafted features \cite{fayad2024hy}. DL models can effectively capture the relationships between input features and canopy height and provide flexibility in network architecture and loss function optimization, which can improve predictions of extreme height values \cite{fayad2024hy}. For instance, \cite{lang2023high} developed a probabilistic DL model that fused sparse GEDI height data with dense Sentinel-2 optical images, producing global canopy height maps at 10-m resolution with a root mean square error (RMSE) of 7.9 m. \cite{fayad2024hy} used a vision transformer (ViT) model with a discrete/continuous loss formulation to map forest canopy height in Ghana, increasing sensitivity to tall trees. Similarly, \cite{liu2023estimation} utilized airborne LiDAR data, 3-m Planet imagery, and a U-Net CNN model to predict tree height and cover in Europe, achieving an average RMSE of 5.4 m. At regional scales, \cite{tolan2024very} employed a self-supervised DINOv2 model to produce 0.59-m resolution canopy height maps for California (US) and São Paulo (Brazil), a significant improvement in spatial resolution over previous studies (\cite{lang2023high}; \cite{potapov2021mapping}). However, this approach relied on airborne LiDAR and high-resolution Maxar imagery, rather than freely available remote sensing data, and required multiple high-end GPUs, which poses a challenge for many research groups \cite{wagner2024sub}. \cite{wagner2024sub} achieved a submeter canopy height for California using a regression U-Net~\cite{ronneberger2015u} CNN model trained on 0.6-m USDA-NAIP imagery and airborne LiDAR, reporting an average RMSE 1.6 times lower than \cite{tolan2024very}. 

The latest studies showed that DL methods associated with high-resolution LiDAR and optical data improved canopy height estimation accuracy (\cite{liu2023estimation, wagner2024sub, tolan2024very}). However, while ALS data availability is restricted to limited regions of the globe, some of the above-mentioned studies use commercial satellite data which are not accessible to some groups, and none of them include the freely available SAR or optical satellite observation features as predictor variables. In addition to making the model more robust when having multiple observation from different seasons, optical data from multiple seasons can provide the model with phenological patterns that can be useful to differentiate forest types and canopy height values. However, optical sensors only measure reflected sunlight from the Earth’s surface which despite showing a correlation to canopy height and consequently to AGB (\cite{barbier2010variation, couteron2005predicting, ploton2013canopy}), tend to saturate at tall canopies, 40 m - 50 m \cite{potapov2021mapping, tolan2024very, wagner2024sub}) or high biomass forests (150-200 Mg/ha; \cite{lu2012aboveground, ploton2013canopy}). Side-looking synthetic aperture radar (SAR), an active sensor that transmits microwaves at a given wavelength and measures the backscatter, can penetrate through forest canopies, with the depth of penetration increasing with increasing wavelengths (\cite{le2011biomass}). In addition, the signal saturation from optical data severely impacts their ability to provide accurate height estimates of dense canopies, especially when using conventional tree ensemble machine learning algorithms, which are unable to learn from the spatial context of the observation at the pixel scale. DL algorithms extract relevant features from raw data as it passes through the network, alleviating the need for hand-crafted features, and have more flexibility in network architecture and loss function optimization (\cite{fayad2024hy}). 

\cite{sothe2022spatially} reported an underestimation of canopy height in Canada. They attributed this to the lack of GEDI observations in that province and many other locations in Canada as GEDI does not cover high latitudes. In this context, Ontario is a province with a considerable area situated south of 51.6° N which made it possible to employ most of the GEDI observations to train the model in \cite{sothe2022spatially} study. Ontario is the second-largest province of Canada with 70.5 million hectares of forest, which represents about 2\% of the world’s forests and 20\% of Canada’s forests (Forest Resources of Ontario, 2021). Ontario’s forest includes temperate, boreal, and tundra forests with distinct coniferous, mixed and deciduous forest regions for which we have large collection of ALS data that can support methodological developments for estimating canopy height for areas outside GEDI coverage. Therefore, it is important to understand how a forest canopy height map could be improved if a model is trained for this province using more complex methods such as DL algorithms that provide spatially continuous uncertainty map and allow for customized weights and make use of spatial features.

Although machine learning models generally exhibit high predictive skill, they often lack spatial uncertainty quantification, which can mask significant inaccuracies in specific areas \cite{duncanson2022aboveground}. Uncertainty maps are critical for decision-making, helping to identify areas where estimates are unreliable and may need to be excluded from analysis,  or treated with particular consideration \cite{lang2022global, lang2023high}. However, relatively few studies provide uncertainty maps alongside canopy height maps, highlighting a gap in the current research that warrants further exploration \cite{lang2022global, lang2023high}.

Here we evaluate the performance of Convolutional Neural Networks with Uncertainty estimates (UCNNs) using the negative log-likelihood of the Laplace distribution to provide spatially continuous canopy height estimates with a measure of uncertainty. Experiments were conducted for 2019 and 2020 in the Province of Ontario, Canada, where temperate, boreal and tundra forests typical of high latitude forests are found. To reduce autocorrelation, we apply spatial cross-validation splitting the study region into five non-overlapping areas. Specifically, we conduct various experiments to investigate (1) the UCNN model performance compared to GEDI observations alone, (2) the benefits of incorporating geographic location and seasonal information from optical and SAR sensors, (3) the characteristics of forested areas with the highest canopy height estimation uncertainty. Through these experiments, we contribute to advancing canopy height estimation methodologies for areas that are not covered by GEDI footprint using advanced machine learning techniques. 





\section{Materials and Methods}

\subsection{Study Area}

\begin{figure}[ht]
    \centering
    \includegraphics[width=0.45\textwidth]{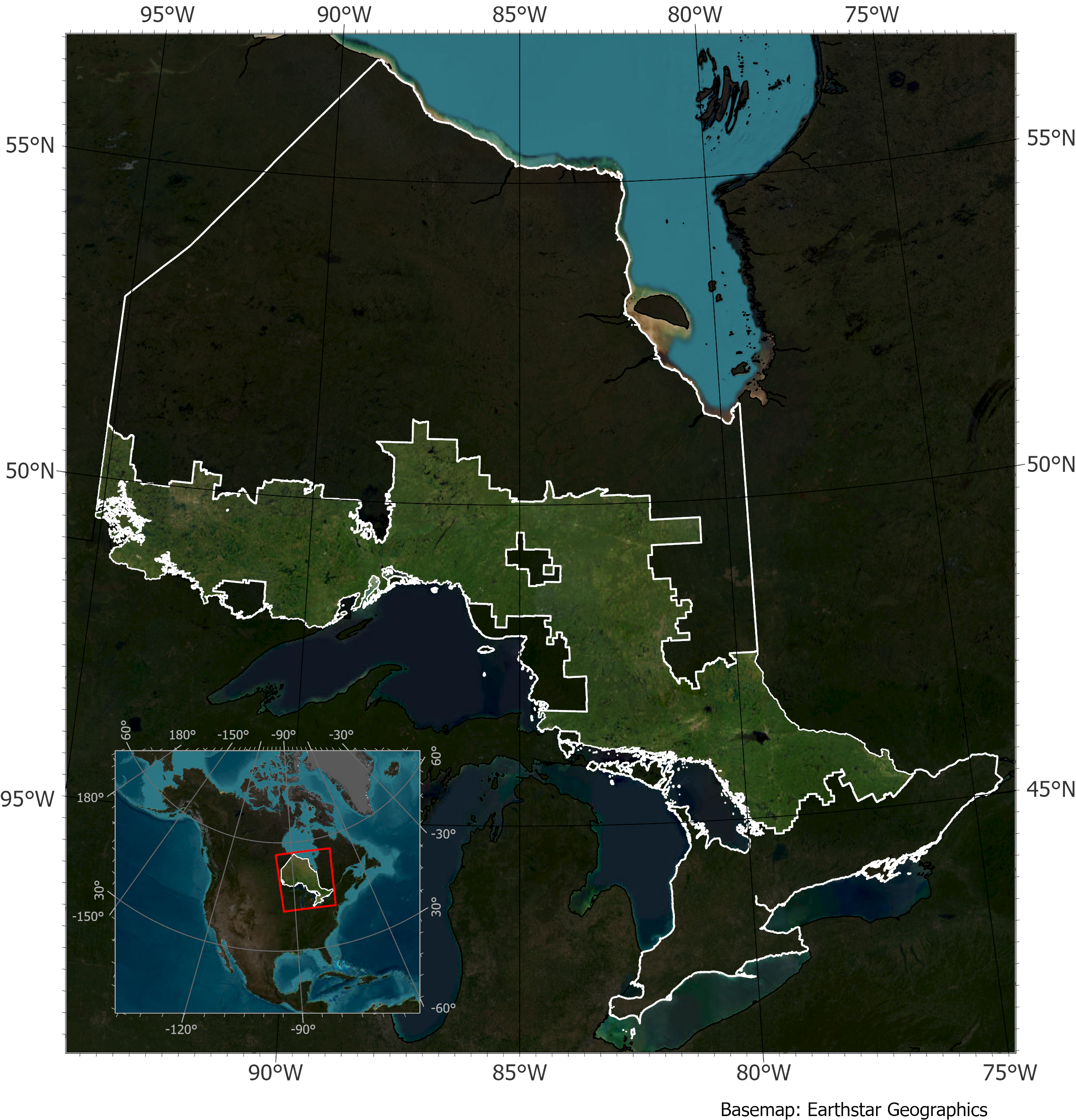}
    \caption{Map of the study area. The highlighted region is the study area for which airborne LiDAR data is available from the Forest Resources Inventory Leaf-on LiDAR dataset of the Government of Ontario.}
    \label{fig:studyarea}
\end{figure}

Experiments were conducted between 2019 and 2020 in Ontario, Canada, the country’s second-largest province, spanning over 1 million square kilometers (1.076 million km²) (see Figure~\ref{fig:studyarea}). Ontario’s climate exhibits a continental nature, characterized by humid conditions in the south, featuring cold winters and warm summers, transitioning to a sub-Arctic in the north. The province’s forests are classified into four primary regions: the Hudson Bay Lowlands in the far north, the boreal forest region, and the Great Lakes–St. Lawrence Forest in the south and central areas, and the deciduous forest in the south encompassing temperate, boreal and tundra forest types of norther high latitudes. The Hudson Bay Lowlands is dominated by stunted Tamarack and Black Spruce well-drained areas with occasional presence of White Birch, Dwarf Birch and Willow. Ontario is part of the Boreal Shield Ecozone which contains conifer species such as Black and White Spruce, Jack Pine, Balsam Fir, Tamarack and eastern White Cedar and few deciduous species like Poplar and White Birch. The Great Lakes–St. Lawrence Forest is dominated by deciduous species such as Maple, Oak, Yellow Birch, White and Red Pine, and mixed forests containing White Pine, Red Pine, Hemlock, White Cedar, Yellow Birch, Sugar and Red Maples, Basswood and Red Oak. The deciduous forest is the southernmost region in Ontario, dominated by agriculture and urban areas with scattered woodlots containing mainly southern deciduous trees and trees that are found in the Great Lakes–St.

\subsection{Data}
\label{sec:data}
\subsubsection{Satellite data and pre-processing}
\label{sec:data_prepro}

In this study, we use spaceborne observations from Landsat 7 and 8, Sentinel-1, ALOS PALSAR-2, and GEDI sensors, all acquired and pre-processed on Google Earth Engine (GEE) platform (Gorelick et al., 2017). Sentinel-1 and ALOS PALSAR operate in the C and L bands, respectively, capturing complementary structural details of the trees, with the first mainly interacting with leaves and top of the canopy while the latter interacts with tree branches and trunk. Meanwhile, Landsat sensors provide spectral information crucial for modeling vegetation dynamics, capturing reflectance information from the forest.  

The datasets were projected to a common EASE-Grid 2.0 North projection with WGS 84 datum at 30-m spatial resolution to align with Landsat pixel size. Our dataset spans from 2019 to 2020, aligning with the availability of ALOS PALSAR-2 and GEDI data availability. For each year, we constructed corresponding seasonal composites for Landsat and Sentinel-1 data, alongside a one-year composite of PALSAR-2. The three monthly median seasonal composites were computed for Winter (January to March), Summer (June to August), and Fall (September to November), for each year to capture the dynamic changes within the study area. Spring data was excluded from our analysis due to substantial spatial variability because of the varying presence of snow between April and May across the study area. We hypothesize that by leveraging these different covariates, including optical, SAR, location, and seasonal information, the neural network will be able to identify changes in tree-related information including structural and spectral aspects of the vegetation. 

To address issues of outliers, noisy data, and gaps in the median seasonal composites for Landsat and Sentinel-1 imagery, we introduce the Seasonal Image Composite Algorithm (SICA), as detailed in Alg.~\ref{alg:composite}, developed on the GEE platform. Inspired by the Multi-year Best Available Pixel (BAP) (Thompson et al., 2015) algorithm, SICA facilitates the generation of smooth image composites for both optical and SAR data by calculating the median value of all valid observations at each pixel location over the analyzed period. In cases where a pixel lacks enough valid observations for a given year and season, SICA employs a recursive approach to retrieve data from the same season in previous years at the same pixel location, as outlined in Alg.~\ref{alg:composite}. This strategy mitigates the impact of data acquisition gaps on computed statistics, ensuring the robustness of the generated image composites. The algorithm is also easily adaptable to various types of sensors, depending on the criteria defined for valid observations. 

For Sentinel 1, we used the Sentinel-1 SAR GRD dataset on GEE. We selected images acquired with the VV (vertical-vertical) and VH (vertical-horizontal) polarizations, interferometric wide (IW) instrument acquisition mode, and ascending orbit pass options. For Landsat, we included Tier 1 surface reflectance images from Collection 2 for the Landsat 7 and 8 instruments available on GEE to increase the data availability in each season, thus increasing the probability of getting cloud-free imagery. We employed a quality filter to remove all pixels flagged as Dilated Cloud, Cirrus, Cloud, or Cloud Shadow based on the QA\_PIXEL bit mask. For Fall and Summer composites, we also excluded snowy pixels from further analysis. Since the available bands differ in the Landsat 7 and 8 surface reflectance products, we selected bands corresponding to blue, green, red, near-infrared (NIR), and shortwave infrared 1 (SWIR 1) and renamed these bands to harmonize the data structure for the three datasets. We then merged the collections into a new harmonized Landsat image collection before applying the recursive compositing scheme. For ALOS PALSAR, we used the associated one-year composite, selecting the HH and HV polarizations.  

From GEDI, we used GEDI L2A relative height 98th (rh98) metric acquired during the summer months (June to August) for each year, filtering out the weak coverage beams and low-quality data using the quality flag information. The rh98 has been used in others research showing high alignment with reference products (Sothe et al. 2022; Duncanson et al. 2022; Liu, Cheng, and Chen 2021). Furthermore, we implemented exclusion criteria for samples exhibiting negative values and those surpassing 40 meters. These criteria were established based on our prior understanding of the forest species and likely tree heights prevalent within the study area. 

\subsubsection{Reference airborne LiDAR (ALS) data}

To validate the accuracy of both GEDI data and the model’s performance, we utilized the Leaf-on airborne LiDAR (ALS) product acquired from the Forest Resources Inventory (FRI) from the Ontario GeoHub data archive (Ontario Ministry of Natural Resources, 2022). This is a high-resolution LiDAR product with a minimum point density of 25 points/m-2 captured via an airborne platform, specifically designed to map various vegetation attributes such as tree cover, density, and height. For our experiments, we retrieved canopy height raster products spanning from 2019 to 2020. The ALS data includes a canopy height map structured into non-overlapping raster tiles, each sized at 1000x1000 pixels with a spatial resolution of 0.5m, resulting in approximately 28,000 tiles per year to cover their study area. Figure~\ref{fig:studyarea} provides an overview of the geographical coverage encompassed by this campaign.

\subsubsection{Comparison with high resolution global canopy height map}

The performance of our models and GEDI data was compared against the performance of META Global canopy height maps ~\cite{tolan2024very} using ALS data as a reference. With 1 m spatial resolution, this dataset offers a comprehensive view of tree canopy height worldwide with eighty percent of the data obtained from imagery acquired between 2018 and 2020. This dataset was generated using DiNOv2 AI model~\cite{oquab2023dinov2}, presenting canopy height with reported mean absolute error of 2.8 m for some reference areas.

\subsection{Methodology}
\label{sec:methodology}

In this study, we estimate spatially continuous high-resolution GEDI canopy height and uncertainty using ensembles of DL regression models. As illustrated in Figure~\ref{fig:methodology}, our approach involves leveraging data from various satellites to train a Fully Convolutional Neural Network (FCN)~\cite{long2015fully} rooted in the ResUnet~\cite{diakogiannis2020resunet} architecture. Specifically, we integrate data from three satellite sources, Sentinel-1, Landsat-7 and -8 sensors, and ALOS PALSAR-2, along with geographic coordinates as covariates to predict the GEDI rh98 canopy height across the study area. 

Our methodology is divided into four stages: data acquisition, data preprocessing, model training, and model inference, as outlined in Figure~\ref{fig:methodology}. The data acquisition stage involves collecting remote sensing data from the aforementioned satellite sources, (Figure~\ref{fig:methodology}A) following the procedure outlined in Section~\ref{sec:data_prepro}. During the data preprocessing (Figure~\ref{fig:methodology}B), to reduce autocorrelation effects, we split the study area into non-overlapping regions, ensuring that the training and testing data are spatially independent. This approach helps to reduce biases in model performance evaluation and enhances the generalization of predictions across varied landscape conditions. Within each region, we generate a dataset comprising coregistered patches, extracted from all covariates and the target variable, using a sliding window approach with overlap. 

During the model training stage, coregistered patches serve as input to the neural network, which is optimized to simultaneously predict both canopy height and the corresponding uncertainty at the pixel level (Figure~\ref{fig:methodology}C). To address different sources of uncertainty, we utilize an ensemble of neural networks, each trained on distinct subsets of the training and validation data from spatially independent regions. In the inference stage (Figure~\ref{fig:methodology}D), this ensemble is applied to generate final predictions for canopy height and associated uncertainty across the test regions. 

The subsequent sections elaborate on the detailed model training procedure, the architecture of the neural network, and the methodology used for quantifying uncertainty during inference: 

\begin{figure*}[!t] 
    \centering 
    \includegraphics[width=1\textwidth]{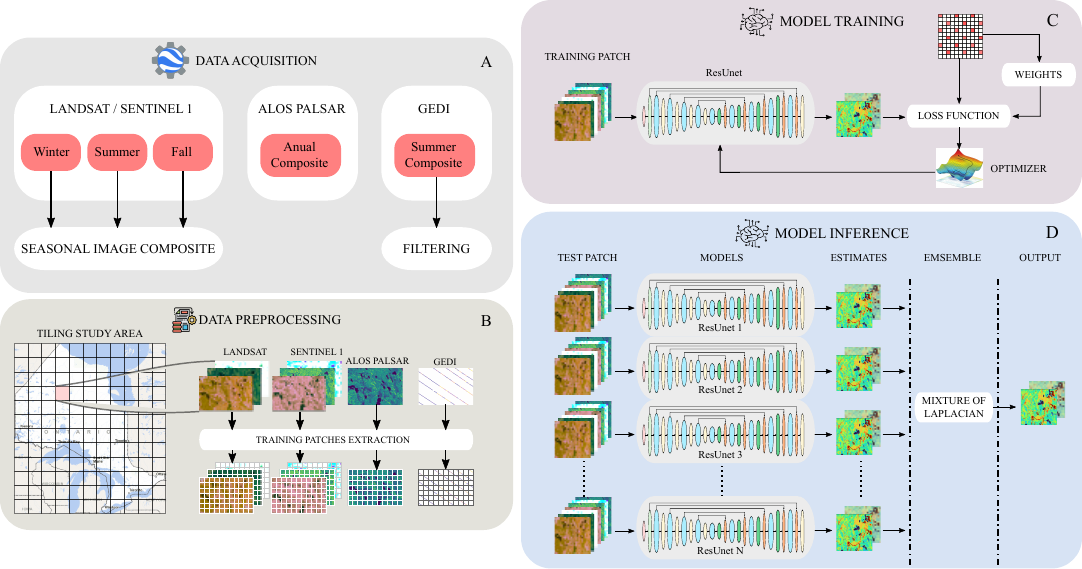}
    \caption{Pipeline of the proposed methodology. a) Descriptions of the data acquisition process. b) Illustration of the data preprocessing strategy followed to extract the training samples. c) Diagram of the neural network optimization procedure phase, and d) illustration of the inference approach using ensemble of neural networks.}
    \label{fig:methodology}
\end{figure*}

\subsubsection{Model Learning}
\label{sec:model_learning}
Let $S=\{x^{(i)}, y^{(i)}\}_{i=0}^N$ represents the set of observed data drawn from an unknown joint probability distribution $J(X,Y)$, where $N$ corresponds to the total number of samples, and the $x^{(i)} \in  X$,  $y^{(i)} \in Y$ pair represents the covariates and the target variable, respectively, the goal of our work is to approximate the underlying conditional probability distribution of $P(Y/X)$, by learning a nonlinear regression function $f_{\theta}: X \rightarrow Y$ through a neural network. In this scheme, each sample $x^{(i)} \in \mathbb{R}^{w \times w \times c}$, $y^{(i)} \in \mathbb{R}^{w \times w \times d}$ pair corresponds to multidimensional coregistered square patches, where $w$ represent the spatial dimensions, $c$ the number of input covariates, and $d$ the number of target variables. 

Similar to (\cite{nair2022maximum}), we optimize the network parameters $\theta$ by minimizing the \textit{Negative Log likelihood} (\textit{NLL}) loss of the \textit{Laplace} distribution:

\begin{equation}
    NNL = \frac{1}{Nw^{2}}\sum_{i=1}^{N}\sum_{j,k \in V^{(i)}}^{N_{V^{(i)}}} \frac{\left| \hat{\mu}^{(i)}_{j,k}-y^{(i)}_{j,k} \right|}{\hat{b}^{(i)}_{j,k}} + \log(2\hat{b}^{(i)}_{j,k})
\end{equation}

where $\hat{\mu}^{(i)}$ and $\hat{b}^{(i)}$ are the neural network expected value and uncertainty estimates, respectively, for a given input sample $x^{(i)}$, where $V^{(i)}$ is a set of valid pixel coordinates in the GEDI map ($y^{(i)}$), and $N_{V^{(i)}}$ represents the associated number of valid pixels. 

In this approach, the network outputs a two-channel map $\hat{y}_{i} \in \mathbb{R}^{h \times w \times 2}$, where the first channel is associated with the expected canopy height (the estimated mean ($\hat{\mu}^{(i)}$)), for each pixel location, and the second one with the corresponding uncertainty ($\hat{b}^{(i)}$) estimates. For a Laplace distribution, the variance is given by $Var=2(\hat{b}^{(i)})^2$~\cite{kotz2012laplace}. During training, because GEDI reference $y^{(i)}$ data is sparse, we masked out pixels without observations during the optimization process using a coordinate matrix $V^{(i)}$, which contains the corresponding valid GEDI observation locations. 
Since GEDI rh98 exhibits a skewed distribution, with a long tail for high canopy height values, analogous to (\cite{yang2021delving, steininger2021density}), we use a weighting function $f_{\omega}(y^{(i)}_{j,k})$ to improve the model's performance across the entire target domain. In particular, $f_{\omega}(y^{(i)}_{j,k})$ adjusts the cost according to the inverse of the kernel density estimate (KDE) of the target variable distribution, where rare height sample values are more heavily penalized in comparison to frequent height values. Then, the loss function becomes,

\begin{equation}
    NNL = \frac{1}{Nw^{2}}\sum_{i=1}^{N}\sum_{j,k \in V^{(i)}}^{N_{V^{(i)}}} f_{\omega}(y^{(i)}_{j,k})\left[ \frac{\left| \hat{\mu}^{(i)}_{j,k}-y^{(i)}_{j,k} \right|}{\hat{b}^{(i)}_{j,k}} + \log(2\hat{b}^{(i)}_{j,k})\right]
\end{equation}


where $f_{\omega}(y^{(i)}_{j,k})$ is given by:

\begin{equation}
    f_{\omega}(z) = \left(\frac{1}{Mh}\sum_{m=1}^{M} K \left(\frac{z -z_{m}}{h}\right)\right)^{-1}
\end{equation}

where $M$, $K$ and $h$ are the total number of data points, a Kernel function and the bandwidth, respectively.

We optimize the neural network using the \textit{NLL} of the \textit{Laplace} distribution instead of the \textit{Normal} because the former exhibits greater robustness to outliers (\cite{nair2022maximum}), thus improving the estimates for both the canopy height mean and uncertainty statistics. In this sense, the \textit{Laplace} distribution is a suitable fit for our data, as we have observed some GEDI samples that display inconsistencies with the intended target, despite filtering for strong and high-quality beams. This phenomenon has been widely documented in other works (\cite{lahssini2022influence,fayad2021quality, tang2023evaluating}), where these errors are generally associated with the effect of weather conditions, such as clouds, transitions between forest and non-forest areas, and geolocation errors.

\subsubsection{Neural Network Architecture}

\begin{figure*}[!t]
    \centering  
    \includegraphics[width=0.7\textwidth]{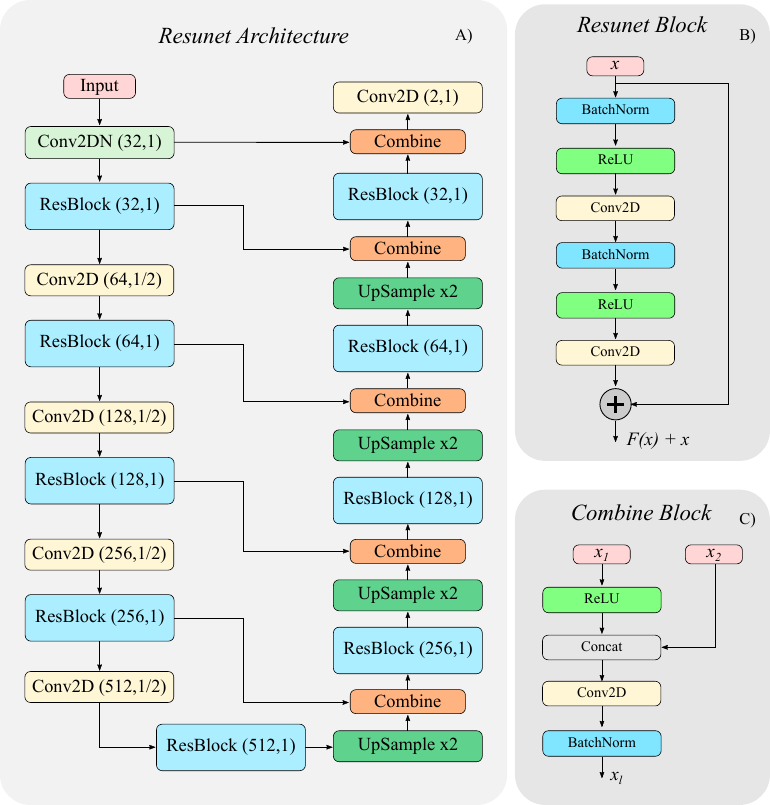}
    \caption{Overview of the ResUNet network. (a) The left (downward) branch is the encoder part of the architecture. The right (upward) branch is the decoder. The last convolutional layer has as many channels as there are distinct classes. (b) Building block of the ResUNet network. Each unit within the residual block has the same number of filters with all other units.}
    \label{fig:network}
\end{figure*}

Figure~\ref{fig:network} illustrates the neural network architecture employed in this study, which is a variation of the ResUNet-a model introduced by~\cite{diakogiannis2020resunet}. Our adaptation involves optimizing the model by reducing the number of parameters and operations. These modifications were implemented to streamline the architecture, thereby improving computational efficiency and speeding up the training process. This adjustment was deemed necessary due to the extensive experimentation conducted in this study. 

In particular, our version of the ResUNet maintains the fundamental structure of the original model, featuring encoder and decoder sections reminiscent of the UNet~\cite{ronneberger2015u} framework. However, instead of using several atrous~\cite{chen2017deeplab} parallel convolution operations in the residual blocks, we opted for traditional residual blocks, as illustrated in Figure~\ref{fig:network}(b). This design choice not only substantially reduces the number of parameters and convolutional operations across the entire network architecture but also makes the network less prone to overfitting. This is particularly significant given the challenges posed by noisy GEDI samples resulting from adverse weather conditions and geolocation errors encountered during the acquisition stage. 

As depicted in Figure~\ref{fig:network}(a), the input data undergoes convolution with a [1, 1] kernel size followed by a Batch Normalization~\cite{ioffe2015batch} layer to increase the number of features to the desired initial filter size. According to~\cite{diakogiannis2020resunet}, the rationale behind choosing a convolutional layer with a [1, 1] kernel size is to prevent potential information loss from the original image. Using a larger kernel size could lead the network to aggregate information from neighboring pixels, potentially resulting in a loss of detail or context. 

Following the initial convolutional operation, the encoder portion of the network takes charge of processing the data. This phase involves a series of steps including residual blocks and downsampling operations. Specifically, downsampling is achieved through convolution with a [1, 1] kernel size and a stride of two, effectively doubling the feature maps. 

Following the encoder stage, the decoder component comes into play to restore the spatial dimensions of the image. This is accomplished through upsampling operations using nearest neighbor interpolation, followed by normed convolution with a kernel size of one. Referred to as Conv2DN, this normed convolution process comprises a single 2D convolution followed by a Batch Normalization layer, aimed at refining the resolution of convolutional features as detailed in~\cite{diakogiannis2020resunet}. 

To integrate the corresponding encoder and decoder feature maps, skip connections are established through the Combine layers. These layers concatenate the two inputs and subject them to normed convolution, adjusting the feature count to the desired size, as illustrated in Figure~\ref{fig:network}(c). Finally, the output of the network is a 2D convolutional layer featuring two outputs with linear activation functions. 

Compared to the original ResUNet-a model, which contains 52 million trainable parameters, our adapted model significantly reduces complexity to 8.8 million parameters. This 5.9-fold reduction not only enhances computational efficiency but also makes the model less prone to overfitting and more robust to noise in the data, thereby improving its generalization capabilities. Additionally, this reduction in model complexity significantly decreases experimental time and hardware requirements, enabling the execution of multiple experiments in a shorter timeframe. 

\subsubsection{Uncertainty Estimation}
\label{sec:uncertainty}
Analogous to~\cite{lang2022global}, our approach employs ensembles of neural networks to characterize both \textit{aleatoric} and \textit{epistemic} uncertainty, as illustrated in Figure~\ref{fig:methodology}. In our methodology, each model in the ensemble outputs a \(Laplace(\hat{\mu},  \hat{b}|x)\) distribution, as described in Section~\ref{sec:model_learning}. Each model is trained to capture and quantify the uncertainty inherent in the data (aleatoric uncertainty) through the $\hat{b}$ estimates. This uncertainty originates from both covariates and GEDI observations, which are affected by factors such as sensor limitations, weather conditions, and geolocation errors. On the other hand, the variations in estimates among the model ensembles for corresponding predictions represent the \textit{epistemic} uncertainty. High concordance among the models indicates lower \textit{epistemic} uncertainty, whereas significant variability suggests higher \textit{epistemic} uncertainty.

Formally, the ensemble's output derived from the mixture of distributions is formulated as:
\begin{equation}
    Y = \sum_{e=1}^{N_e} \pi_{e} Laplace(\hat{\mu}_{e},  \hat{b}_{e}|x)
\end{equation}
where $\pi_{e}$, $e = 1,2,...,N_e$ are the mixing probabilities that satisfy $\pi_{e} \geq 0 $ and $\sum_{e=1}^{N_e} \pi_{e} = 1 $, and $N_e$ indicates the number of model constituting the ensemble.

The total mean (first moment) and total variance  (second moment) are given by:

\begin{equation}
    E[Y] = \mu =  \sum_{e=1}^{N_e} \pi_{e} \hat{\mu}_{e}
\end{equation}

\begin{equation}
    \begin{aligned}
    E[(Y-\mu)^2] & = \sigma^2 =  E[Y^2] - (E[Y])^2 \\
    & = \left( \sum_{e=1}^{N_e} \pi_{e} (E[Y_{e}^2]) \right) -\left(\sum_{e=1}^{N_e} \pi_{e} \hat{\mu}_{e}\right)^2 \\
    & = \sum_{e=1}^{N_e} \pi_{e} (\hat{\sigma}_{e}^2 + \hat{\mu}_{e}^2) -\left(\sum_{e=1}^{N_e} \pi_{e} \hat{\mu}_{e}\right)^2 \\
    & = \sum_{e=1}^{N_e} \pi_{e} (\hat{\sigma}_{e}^2 + \hat{\mu}_{e}^2) -\left(\sum_{e=1}^{N_e} \pi_{e} \hat{\mu}_{e}\right)^2 \\
    & = \sum_{e=1}^{N_e} \pi_{e} \hat{\sigma}_{e}^2 + \sum_{e=1}^{N_e} \pi_{e} \hat{\mu}_{e}^2 -\left(\sum_{e=1}^{E} \pi_{e} \hat{\mu}_{e}\right)^2
    \end{aligned}
\end{equation}

Assuming equally probability $\pi_{e}=1/N_e$ and Laplace variance $\hat{\sigma}_{e}^2=2\hat{b}_{e}^2$, we have: 

\begin{equation}
    \label{eq:total_variance}
    E[(Y-\mu)^2] = \left[ \frac{2}{N_e}\sum_{e=1}^{N_e} \hat{b}_{e}^2 \right] + \left[ \frac{1}{N_e} \sum_{e=1}^{N_e} \hat{\mu}_{e}^2 -\left(\frac{1}{N_e}\sum_{e=1}^{N_e} \hat{\mu}_{e}\right)^2\right]
\end{equation}
where the first term in Eq.~\ref{eq:total_variance} corresponds to the \textit{aleatoric uncertainty} and the second's one to the \textit{epistemic uncertainty}.

While the adopted strategy allows us to quantify the total uncertainty of predictions, this does not necessarily indicate that the estimated uncertainty is well-calibrated~\cite{laves2019well, laves2020well, kuleshov2018accurate, dawood2023uncertainty}. Calibration in this context refers to the alignment between the model’s predicted uncertainties and their actual reliability~\cite{zhang2023study}.. A well-calibrated model not only provides uncertainty estimates but also ensures that these estimates accurately reflect the true variability in predictions. Ideally, a well-calibrated model exhibits a linear relationship between predicted uncertainty and actual prediction errors~\cite{laves2019well, laves2020well, kuleshov2018accurate}. This means that for a given uncertainty level, the average error of the model’s predictions should correspond closely to the predicted uncertainty. Achieving calibration is crucial for ensuring the reliability and usefulness of uncertainty estimates in making informed decisions based on model predictions. 

Among the different approaches to calibrate models, we adopted the Platt-based~\cite{platt1999probabilistic} scaling strategy, adapted for regression problems in~\cite{laves2019well, laves2020well, kuleshov2018accurate}because it does not modify the predictive performance of the model. Instead, it only adjusts the uncertainty outcomes to ensure they are well-calibrated. This approach allows the model to maintain its accuracy while providing reliable uncertainty estimates, ensuring they accurately represent the confidence in the predictions. Notice that we apply this procedure after the ensemble estimates, so that we modify the total uncertainty coming from the mixture of Laplace distributions. 

Formally, a well-calibrated estimation of predictive uncertainty for a regression model is defined by:

\begin{equation}
\begin{aligned}
    E_{x,y} = \left[ \lVert y - \hat{y}\rVert^2 | \hat{\Sigma}^2=\Sigma^2  \right]=\Sigma^2,  \forall \left\{ \Sigma^2 \in R| \Sigma^2 \ge 0\right\}
\end{aligned}
\label{eq:calibration}
\end{equation}
where $\Sigma^2$ represents the total uncertainty. To calibrate the model uncertainty ($\hat{\Sigma}^2$), an optimization process is carried out.  This process involves the use of a validation set to determine a scalar $s$, which acts as a corrective factor. The scalar $s$ is optimized through an iterative process, where the objective is to minimize the difference between the predicted uncertainties and the actual observed errors, adjusting the $\hat{\Sigma}^2$ outcomes to approximate the definition given in Eq.~\ref{eq:calibration}. For more details, see~\cite{laves2019well, laves2020well, kuleshov2018accurate}.

\subsubsection{Harmonize GEDI and External Datasets}
\label{sec:als_harmonizing}

\begin{figure*}[!th]
    \centering    \includegraphics[width=0.7\textwidth]{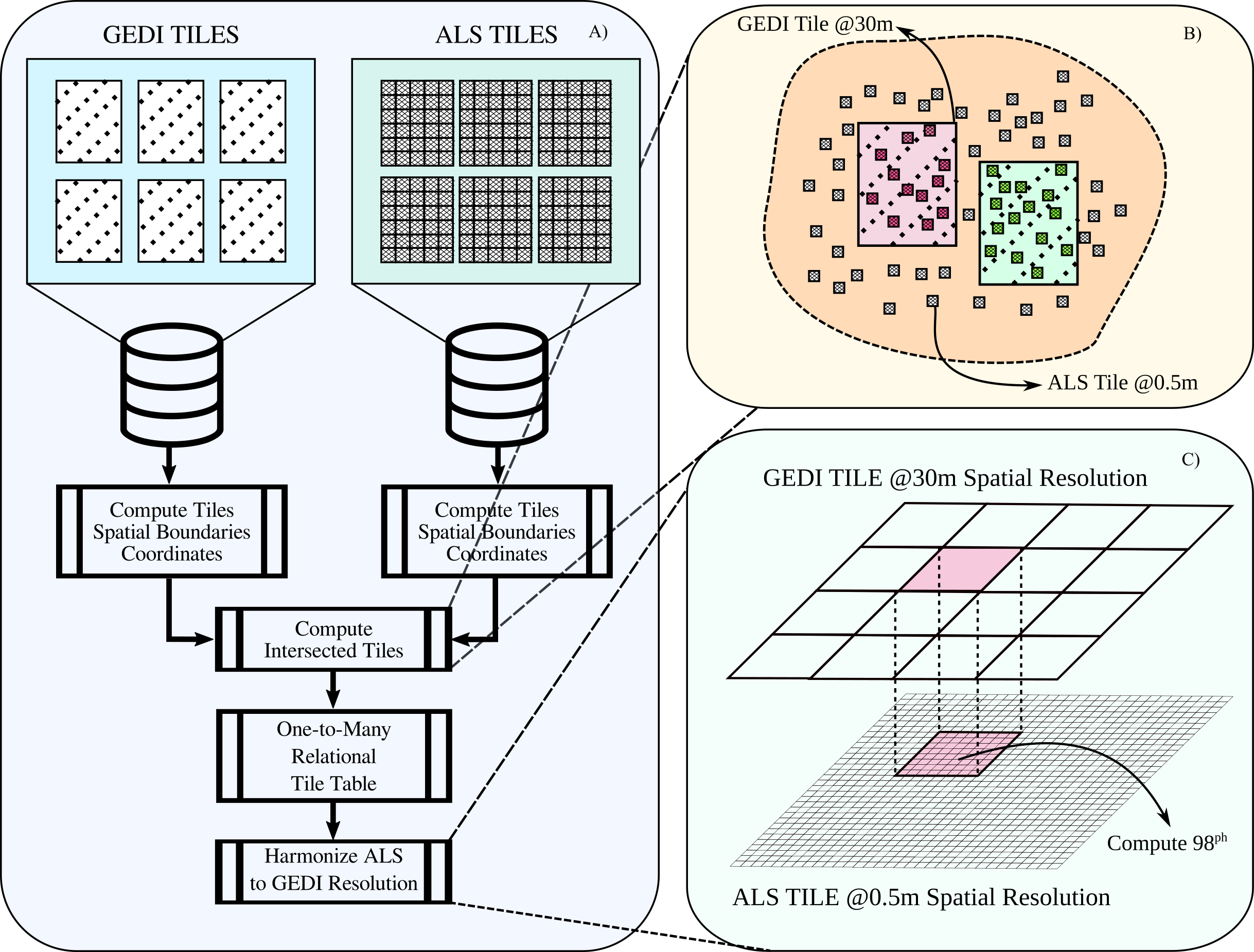}
    \caption{GEDI and ALS harmonization methodology. a) pipeline of the harmonization strategy. b) Illustration of the GEDI and ALS interception tiles. c) Zonal statistics energy-based process to harmonize ALS samples with corresponding GEDI.}
    \label{fig:harmonize}
\end{figure*}

Figure~\ref{fig:harmonize} illustrates the pipeline utilized to harmonize ALS and META data with corresponding GEDI measurements, addressing the difference in spatial resolution between GEDI and the ALS and META data. The primary objective is to generate ALS and META products at the same pixel spatial resolution as GEDI. This harmonization allows for the validation of both GEDI observations and model predictions against the ALS reference data, as well as the META maps. 

As described in Figure~\ref{fig:harmonize}(a), the process begins with the computation of spatial boundary coordinates for both GEDI and high-resolution tiles (ALS and META tiles), using the same coordinate system. Then, for each GEDI tile, we determine the intersecting high-resolution tiles, as shown in Figure~\ref{fig:harmonize}(b). This relationship is structured in a One-to-Many Relational Table, linking each GEDI tile with all high-resolution tiles related to the same geographic location. 

Finally, high-resolution data is harmonized with GEDI spatial resolution using a zonal statistics algorithm, which computes statistics of high-resolution pixel products overlapping with each GEDI pixel, as depicted in Figure~\ref{fig:harmonize}(c), where the unique GEDI tiles delineate each zone. Specifically, for a given GEDI tile, the spatial boundary coordinates of all its pixels are first estimated. Then, for each pixel extent, high-resolution pixels overlapping with it are identified, and the 98th percentile of these high-resolution pixels is computed. This process generates ALS and META products at GEDI resolution, aligning with the GEDI acquisition energy-based scheme for the rh98 product, which measures the accumulated reflected energy at the observed location in percentiles.

\subsection{Experimental Design}

The study region was partitioned into a grid of 100 non-overlapping tiles, each spanning approximately 8000x7000 pixels (Figure~\ref{fig:methodology}). Spatial cross-validation was conducted using five-folds, with 80\% of the tiles designated for the training set and 20\% for the testing set in each fold. Within each fold, the training set was further divided into five spatially random sub-folds. Subsequently, models were trained using five different partitions of training and validation data, ensuring each model utilized a distinct training/validation setup. This strategy helped avoid bias towards a particular training/validation configuration during the model optimization process, thereby capturing the statistical variability across the study area. Furthermore, this approach allowed for the assessment of the models’ generalization capabilities across diverse forest types, land cover, and land use conditions while mitigating bias arising from inherent spatial autocorrelation. Consistency in the partition configuration was maintained throughout all experiments to ensure a fair comparison of models under similar conditions. 

During the inference stage, the five trained models corresponding to each fold were executed in the associated testing areas. The outcome for each pixel was determined using a model ensemble applying the mixture Laplace distribution, as described in Section~\ref{sec:uncertainty}. 

To reduce the potential impact of artifacts during the training process, extreme values for each covariate were normalized to ensure their adherence to the intervals ($\mu - 3\sigma$, $\mu + 3\sigma$), where $\mu$ and $\sigma$ denote the corresponding covariate empirical mean and standard deviation, respectively. Given the extensive dataset required to encompass the entire study area, we adopted an online computation strategy for calculating the statistics of each covariate. This process was carried out using the \cite{chan1983algorithms} algorithm implemented in \cite{pedregosa2011scikit}. Then, to facilitate the network optimization process the standard normalization was applied using computed statistics~\cite{huang2023normalization}.

To generate the training samples, we systematically extracted coregistered patches of 64x64 pixels from all sensors for each year and tile. This was achieved using a sliding windows procedure with a 25\% overlap across all tiles, focusing on areas with GEDI observations. We employed this patch size because it represents the smallest area with higher GEDI pixel density per meter square, according to its sampling pattern described in section~\ref{sec:data}. We excluded patches containing GEDI pixels less than 1\% of the patch size to reduce the impact of sparse samples during the optimization process. Furthermore, by implementing an overlapping sampling strategy, we circumvented the issue of missing locations. Finally, the extracted samples were exported into TFRecords files for efficient I/O data transferring during training~\cite{tensorflow2015-whitepaper}. 

During training, we computed the GEDI density sample distribution at each run, randomly sampling 10 million samples from the training set. Then, we estimated the inverse distribution of weights for the weighted loss function, following the methodology described in section~\ref{sec:methodology}. Neural Network weights were randomly initialized using the truncated normal distribution HeNormal. During training, weights were also regularized using the $L2$ weight decay strategy to reduce overfitting, which parameter was set to 10e-5. We employed the Adam~\cite{kingma2014adam} optimizer to learn the neural network’s parameters, setting the learning rate to 1e-4 with a Cosine schedule~\cite{loshchilov2016sgdr}, and clipping the gradients to norm 1.0. We set the batch size to 256, and epochs to 50.

\subsubsection{Features}

We conducted extensive experiments to investigate the influence of various factors on model performance, focusing particularly on seasonal, sensor, and geographic coordinates covariates (Table~\ref{tab:results}). Our experimental approach involved systematically permuting all combinations of these covariates to uncover patterns and elucidate their impact on model outcomes. 

In our analysis of seasonal information, we conducted experiments using both the entire seasonal dataset and separate experiments focusing exclusively on data from the summer months. This enabled us to evaluate the contribution of seasonal information to the learning process. We hypothesized that during the winter months, the presence of snow would delineate image structural information, thereby enhancing SAR-based sensors’ ability to capture intricate details of the study area’s topography. Conversely, during the summer and fall months, we anticipated that spectral reflectance would play a more significant role in model performance due to the varying spectral signatures exhibited by different tree and plant species during this time of the year. By leveraging both sources of features, and the differences in summer and winter feature values, we expected to achieve improved modeling of canopy height estimates. 

In our sensor analysis, we aimed to assess the significance of L-band data from the one-year ALOS-PALSAR composite. L-band waves possess the capability to penetrate through the dense canopy of trees, reaching down to the stems and capturing valuable information about the internal structure of the vegetation. This data is crucial for understanding forest dynamics and accurately estimating canopy height. By complementing the canopy-surface information captured by the Sentinel-1 sensor’s C-band information, we anticipated that the inclusion of L-band data would enhance our ability to analyze the entire vertical profile of the forest canopy. 

Furthermore, we conducted experiments to evaluate the impact of geographic location on the model’s generalization capabilities across the study area. We aimed to uncover any spatial biases in the model predictions. This analysis enabled us to identify regions where the model may struggle to accurately estimate canopy height and ensure robust performance across diverse landscapes and environmental conditions present within the study area.

\subsubsection{Accuracy Assessment}
\label{sec:accuracy_assessment}

To evaluate the accuracy of canopy height estimation, we conducted a comparative analysis using data from GEDI, Harmonized META, and trained DL models, comparing them against Harmonized ALS data, generated using the methodology detailed in section~\ref{sec:als_harmonizing}. Specifically, we collected GEDI samples spatially intersecting with the corresponding Harmonized ALS product for each year and tile. These samples were then compared with the dense maps generated by the DL models, as well META maps, at the corresponding geographic coordinates. It is important to note that our evaluation encompasses data from both 2019 and 2020, allowing for a comprehensive assessment that includes more samples containing GEDI data overlaying the study area and ALS campaign. This expanded dataset enables a more robust evaluation of the performance statistics. 

We quantify the performance of evaluated estimates using the coefficient of determination (R-square or R\textsuperscript{2}), the root mean square error (RMSE), the mean absolute error (MAE), and the estimated Bias, metrics commonly used in the context of regression analysis. In addition to numerical metrics, scatter plots serve as valuable tools for visually assessing the relationship between predicted and actual values. These plots offer insights into patterns, trends, and potential outliers in the data, complementing the information provided by numerical metrics. Furthermore, a visual inspection analysis is conducted by comparing generated maps with the reference Harmonized ALS data. Uncertainty maps are examined to discern patterns across different land cover areas, particularly where DL models may encounter challenges in accurately characterizing the mapping process. By leveraging these metrics and visualizations, we can conduct a comprehensive evaluation of the performance of our regression models in estimating canopy height using GEDI data and DL model predictions.

\section{Results}

\subsection{Accuracy Analysis}

Table~\ref{tab:results}  summarizes the performance of each experimental results in terms of R\textsuperscript{2}, RMSE, MAE, and Bias. These results were computed after filtering out canopy height values lower than 3 meters and higher than 40 meters for both ALS and META products, ensuring fair comparisons. The table is organized as follows: the first five rows, from top to bottom, detail the performance of experiments conducted using all available seasonal data, with permutations applied to various covariates. The subsequent five rows show the corresponding results for experiments that utilized only summer data. The final two rows serve as reference benchmarks, presenting the performance of GEDI observations against Harmonized ALS data and the META canopy height maps harmonized to match GEDI's spatial resolution. 

\begin{table}
\centering
\caption{Summary performance statistics of estimated canopy heights with various sensors and experiments in terms of R\textsuperscript{2}, Mean Absolute Error (MAE),  Root Mean Square Error (RMSE) and estimated Bias. W=Winter, S=Summer, F=Fall, for seasons, and LS=Landsat, S1=Sentinel-1, AP=ALOS PALSAR-2, for sensors, LL=Latitude and Longitude coordinates, and X=Omitted from the experiment. For all statistics, Harmonized ALS data were employed as reference data. }
\label{tab:results}
\begin{tabular}{|c|c|cccc|}
\hline
\multirow{2}{*}{\textbf{Seasons}} & \multirow{2}{*}{\textbf{Sensors}} & \multicolumn{4}{c|}{\textbf{Metrics}}                                                                                     \\ \cline{3-6} 
                                  &                                   & \multicolumn{1}{c|}{\textbf{R2}} & \multicolumn{1}{c|}{\textbf{MAE}} & \multicolumn{1}{c|}{\textbf{RMSE}} & \textbf{Bias} \\ \hline
W-S-F                             & LS-S1-AP-LL                       & \multicolumn{1}{c|}{0.72}        & \multicolumn{1}{c|}{3.43}         & \multicolumn{1}{c|}{2.56}          & 2.44          \\ \hline
W-S-F                             & LS-S1-AP-X                        & \multicolumn{1}{c|}{0.69}        & \multicolumn{1}{c|}{3.60}         & \multicolumn{1}{c|}{2.64}          & 3.10          \\ \hline
W-S-F                             & LS-S1-X-LL                        & \multicolumn{1}{c|}{0.69}        & \multicolumn{1}{c|}{3.62}         & \multicolumn{1}{c|}{2.68}          & 3.01          \\ \hline
W-S-F                             & LS-S1-X-X                         & \multicolumn{1}{c|}{0.72}        & \multicolumn{1}{c|}{3.46}         & \multicolumn{1}{c|}{2.60}          & 3.49          \\ \hline
W-S-F                             & LS-X-X-X                          & \multicolumn{1}{c|}{0.63}        & \multicolumn{1}{c|}{3.95}         & \multicolumn{1}{c|}{2.95}          & 3.04          \\ \hline
X-S-X                             & LS-S1-AP-LL                       & \multicolumn{1}{c|}{0.62}        & \multicolumn{1}{c|}{3.98}         & \multicolumn{1}{c|}{2.97}          & 4.38          \\ \hline
X-S-X                             & LS-S1-AP-X                        & \multicolumn{1}{c|}{0.58}        & \multicolumn{1}{c|}{4.22}         & \multicolumn{1}{c|}{3.2}           & 3.44          \\ \hline
X-S-X                             & LS-S1-X-LL                        & \multicolumn{1}{c|}{0.55}        & \multicolumn{1}{c|}{4.36}         & \multicolumn{1}{c|}{3.29}          & 4.69          \\ \hline
X-S-X                             & LS-S1-X-X                         & \multicolumn{1}{c|}{0.60}        & \multicolumn{1}{c|}{4.11}         & \multicolumn{1}{c|}{3.09}          & 3.18          \\ \hline
X-S-X                             & LS-X-X-X                          & \multicolumn{1}{c|}{0.51}        & \multicolumn{1}{c|}{4.55}         & \multicolumn{1}{c|}{3.43}          & 4.56          \\ \hline
                                  & \textbf{Products}                 & \multicolumn{4}{c|}{}                                                                                                     \\ \hline
S                                 & GEDI                              & \multicolumn{1}{c|}{0.58}        & \multicolumn{1}{c|}{4.18}         & \multicolumn{1}{c|}{2.73}          & -0.49         \\ \hline
-                                 & META                              & \multicolumn{1}{c|}{0.11}        & \multicolumn{1}{c|}{6.85}         & \multicolumn{1}{c|}{5.86}          & -3.1          \\ \hline
\end{tabular}
\end{table}

The experiments conducted to validate the GEDI product by harmonizing ALS data with GEDI spatial resolution indicate that the GEDI sensor captures 58\% of the variability in the ALS data, with a RMSE close to 4 meters and a (MAE of less than 3 meters. In contrast, the validation of the META product, also harmonized with ALS data, shows that it captures only 10\% of the reference data's variability, representing a 48\% reduction compared to the GEDI product. Additionally, the RMSE and MAE for the META product increase to 6.85 meters and 5.86 meters, respectively.  

By analyzing the results of our models, the best outcomes for the evaluated models occurred when incorporating all seasonal and sensor data covariates. For these experiments, the model estimates achieved 0.72, 3.43 m, and 2.56 m of R\textsuperscript{2}, RMSE, and MAE, respectively. Notably, this model not only outperformed other models but also exhibited superior performance compared to the results obtained from GEDI data alone, demonstrating higher R2 and lower errors. This behavior is consistent across various experimental configurations using seasonal data, where neural network estimates consistently surpassed our benchmark in terms of R\textsuperscript{2} and RMSE.  

Among the experiments conducted using seasonal data, the performance was notably lower when only optical data was utilized. In this configuration, the R2 decreased significantly from 0.72 to 0.63. Moreover, the RMSE increased from 3.43 m to 3.95 m, and the MAE increased from 2.56 m to 2.95 m. In particular, we can notice in Table~\ref{tab:results} that the exclusion of Sentinel 1 information has more impact in model accuracy than the exclusion of ALOS-PALSAR-2 data and geographic information. In fact, the experiments employing Landsat and Sentinel 1 data alone presented lower reduction in performance compared to using all covariates.  

Consistently, the worst results were obtained when only summer and optical information were used, i.e., excluding both winter and fall seasonal data, as well as Sentinel 1, ALOS-PALSAR-2, and geographic information. Notice that experiments exclusively using summer data exhibited a consistent decline in performance across all metrics compared to the corresponding ones that incorporate all seasonal information.  

Regarding our ALOS-PALSAR-2 experiments, superior performance was consistently observed across all evaluated metrics when complementing the L-band data with Landsat and Sentinel-1 C-band covariates. Notably, the impact of ALOS-PALSAR-2 data was particularly evident in experiments using only summer data, which showed a reduction in on the R2, and an increase of RMSE and MAE metrics when compared to the experiments using ALOS-PALSAR information. 

\begin{figure*}[!th]
    \centering
    \includegraphics[width=1.0\textwidth]{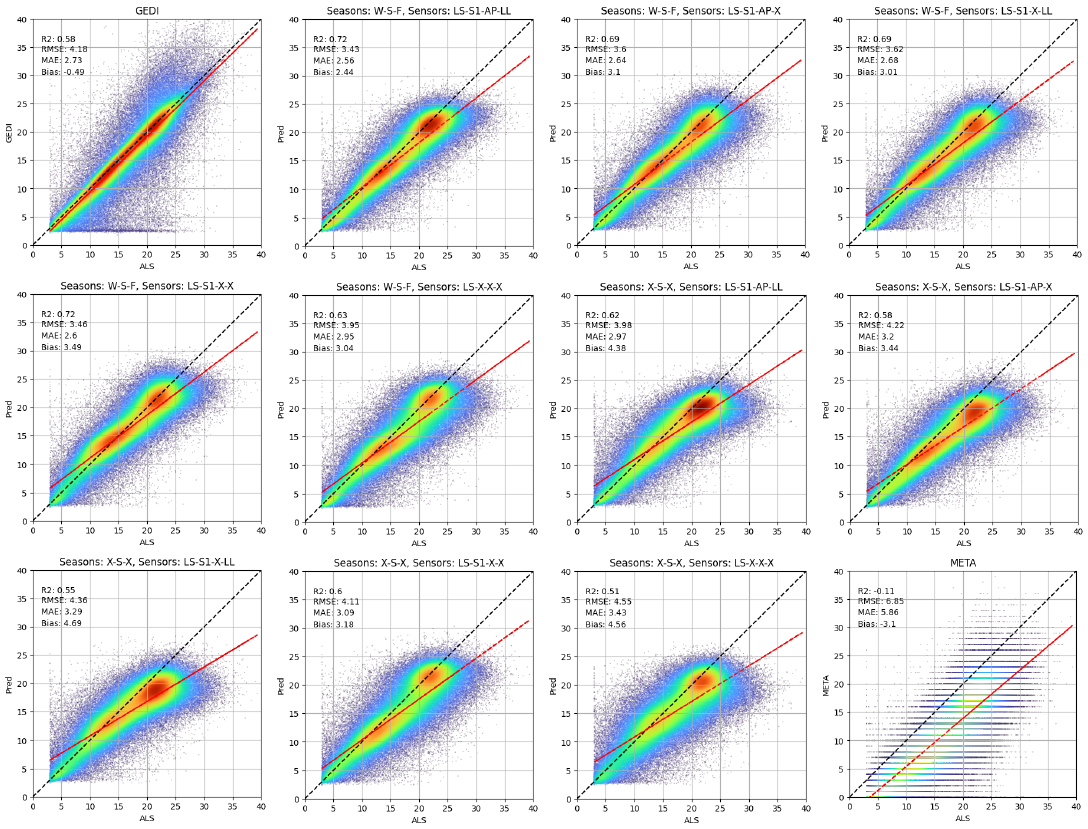}
    \caption{Relationship of canopy height from GEDI, this study and META with harmonized ALS reference data. The 1:1 line between observed and predicted values is shown in dashed black line and the estimated regression in bold red line. Red color density plot indicates a high concentration of samples, while green and blue colors represent medium and low-density ranges, respectively.}
    \label{fig:scatter_plots}
\end{figure*}

\begin{figure*}[!t]
    \centering  
    \includegraphics[width=1\textwidth]{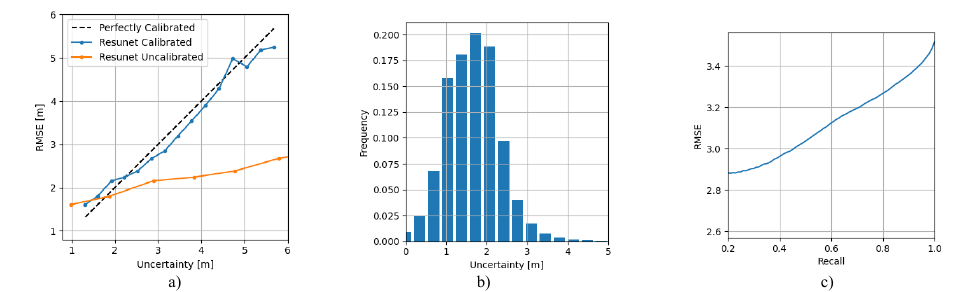}
    \caption{Uncertainty calibration results. (a) Uncertainty calibration curves before and after calibrated. (b) Distribution of samples uncertainties in meters after the model calibration. (c) Precision Root mean square error (RMSE) vs Recall after sorting samples based on the uncertainty measure}
    \label{fig:calibration}
\end{figure*}

Figure~\ref{fig:scatter_plots} presents the density scatterplots illustrating the relationship between GEDI observations, META estimates, and the outcomes of all evaluated model configurations against Harmonized ALS data. Across various model configurations, a high correlation between predictions and reference data can be observed. The density scatter plots show that most data points cluster around the ideal linear regression line, indicating robust alignment between predicted values and ground truth across different model setups. Comparing the scatterplots of GEDI observations with those of the model predictions reveals that GEDI exhibits a stronger linear relationship with the reference data across the evaluated sample domain. Specifically, the density of GEDI observations appears narrower compared to the density of the model predictions, with bias close to zero in the estimated regression line. Additionally, across the entire target domain, GEDI provides more balanced predictions, indicating that GEDI predictions are more evenly distributed rather than concentrated in a specific range. This behavior also validates the methodology used to harmonize the ALS data, which aimed to mimic the GEDI energy reflection-based scheme. 

However, it is also apparent that GEDI presents more outliers compared to some models (Figure~\ref{fig:scatter_plots}). A detailed analysis of the GEDI scatter plots shows that for low-range ALS reference canopy height values, some GEDI samples report high canopy height values. In some instances, these samples can exceed 30 meters when the reference is less than 5 meters. Similarly, for certain high-range ALS canopy height values, GEDI observations indicate low canopy height values.  

Analyzing the scatterplots of the models’ predictions reveals that, in most cases, the models’ predictions are concentrated around the ideal regression line with a less narrow sample density compared to GEDI, with higher bias, but with fewer outliers. For high-range ALS reference values, most models corrected the GEDI estimates effectively, resulting in a lower concentration of outliers. Similarly, for low-range ALS values, the models corrected some outliers, although a few persist in these ranges. These observations align with the global metrics, where in some experiments, the trained models outperformed the GEDI observations.  

Regarding the difference between scatterplots of models trained with seasonal and only summer information, there’s an observed increase in outliers in models trained solely with summer data and a reduced capacity to estimate high target values. Seasonally trained models demonstrate more consistent predictions, especially for high canopy height values, as depicted in the density scatterplots. 

Another notable behavior is that models trained using only summer data exhibit scatterplot densities with less concentration around the ideal regression line. Additionally, it is evident that these models exhibit higher bias compared to those trained with seasonal information. Specifically, the bias for models using seasonal data ranges from 2.4 to 3.1 m, whereas for models trained with summer data, the bias ranges from 3.5 to 4.7 m in the worst-case scenario.  

There is a noticeable lower correlation between META canopy height and the reference ALS data, as summarized in Table~\ref{tab:results} and Figure~\ref{fig:scatter_plots}. Specifically, the density of samples is indicated by blue-green colors, which show less concentration around the estimated regression line. Additionally, the META data is more dispersed, with numerous outliers across the entire domain and a bias towards lower values. 

\subsection{Uncertainty Analysis}

Figure~\ref{fig:calibration} (a) shows the uncertainty calibration curves of the model with the best performance, specifically the model trained using seasonal information and all covariates. We followed this strategy, considering that the calibration methodology we adopted does not affect the model’s predictive performance. Instead, the calibration process solely modifies the model’s uncertainty estimates. For this calibration, we adjusted the total uncertainty estimate, which is derived from aggregating both aleatory and epistemic uncertainty. In this figure, the model’s uncertainty calibration curve is displayed both before and after the calibration, demonstrating how the calibration process improves the alignment of the estimated uncertainties with the RMSE. 

Notably, before calibration, the uncertainty estimates were not meaningful for making decisions about the operational set point based on the RMSE. Under these conditions, as the uncertainty increased, the RMSE did not increase proportionally. After performing the calibration process, the uncertainty curve aligned more closely with the ideal calibration line. With the calibrated model, we can now indicate the operational range of the model where we are certain that, on average, the model will have that prediction’s error in terms of RMSE. 

Complementary information about the uncertainty calibrated model is presented in Figure~\ref{fig:calibration} (b) and Figure~\ref{fig:calibration} (c), corresponding to distributions of calibrated uncertainties estimates, and the RMSE vs recall curve, respectively. By analyzing Figure~\ref{fig:calibration} (b), it can be observed that most of the uncertainties are concentrated between one and three meters. 

Similarly, the precision (RMSE) vs. recall curve indicates that the RMSE of most of the samples is concentrated at values lower than 3.4 m. This curve was generated by sorting the samples in increasing order according to their uncertainty estimates. Then, the RMSE was computed for the progressively increasing proportion of samples until reaching 100\% of the evaluated samples. This approach demonstrates how the model’s performance varies with the level of uncertainty, highlighting its reliability in making predictions within specified uncertainty bounds.

\subsection{Visual Analysis}

\begin{figure*}[t]
    \centering
    \includegraphics[width=1.0\textwidth]{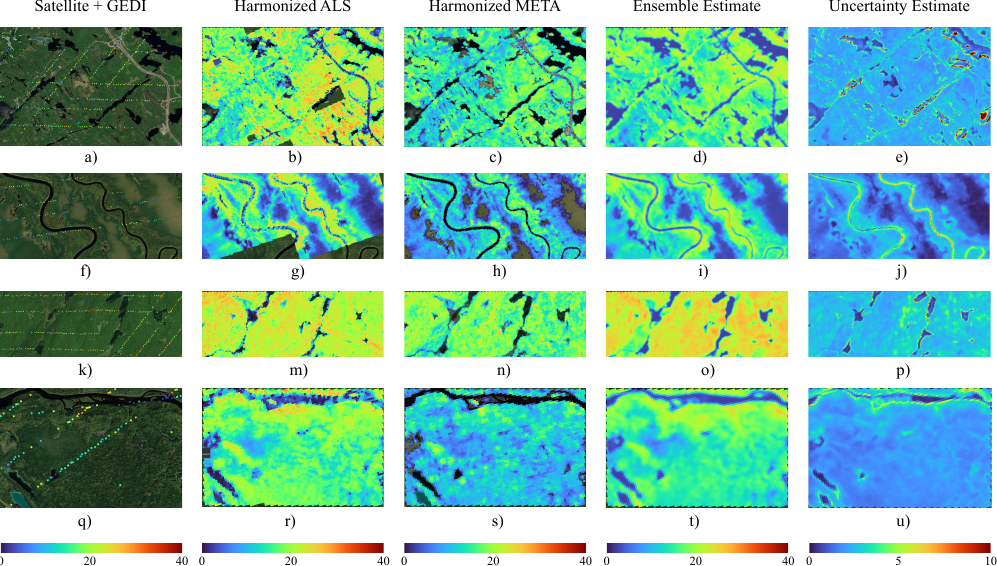}
    \caption{Snapshots of three locations where GEDI and ALS canopy height data intersect. From left to right, the first column shows the GEDI observations overlaid on satellite image, the second column shows the Harmonized ALS maps, the third column presents the Harmonized META maps, the fourth column shows the predicted maps using the best model, and the last column shows the corresponding calibrated uncertainty maps. Dark blue colors represent lower targets, green and yellow signify mid-height targets, and red colors indicate higher height values.}
    \label{fig:snips}
\end{figure*}

Figure~\ref{fig:snips} shows locations within the study area that have both GEDI and ALS data. Note that the selected regions show transitions between forest and non-forest areas, and water bodies. The CH maps, based on their color codification, reveal distinct variations in canopy height targets across the depicted areas, effectively covering the entire range of heights. In the first column, the sparse sampling pattern of the GEDI acquisition campaign can be seen. Consequently, many areas remain unobserved, leading to gaps in capturing important land cover characteristics within the study area. Furthermore, some points along the GEDI stream are missing due to the filtering process applied to consider only strong and high-quality beams. As a result, certain areas, such as water bodies, may lack valid GEDI observations. 

Furthermore, when comparing the snapshots of Harmonized ALS data with the Google satellite base image, a positive correlation is observed across all land surfaces. For instance, water bodies are depicted with zero height, while areas showing signs of deforestation are represented in the ALS maps with lower values. Higher values correspond to regions with dense forest cover. This analysis reinforces that the harmonization process described in Section~\ref{sec:als_harmonizing} maintains the land cover characteristics of the study area. Additionally, it is evident that the ALS campaign does not cover the entire study area; there are regions with missing observations, as shown in Figure~\ref{fig:snips}~(g). 

Similarly, the Harmonized META maps exhibit patterns comparable to both the satellite base image and Harmonized ALS data, effectively delineating water bodies and distinguishing between forested and non-forested land covers. However, a noticeable trend is that the META maps consistently show lower canopy height estimates compared to the reference ALS data, especially in forested areas with high target values. This pattern aligns with the bias statistics, which reveal a significant negative bias in the META product. In these locations, the ALS maps display yellow-red colors, indicating higher canopy height values, whereas the corresponding META maps predominantly underestimate these values. This behavior is particularly evident when comparing Figure~\ref{fig:snips}~(m) and Figure~\ref{fig:snips}~(n) and Figure~\ref{fig:snips}~(r), and Figure~\ref{fig:snips}~(s) snapshots. These findings align with the scatterplot analysis, which indicated that META estimates are biased towards lower values, thus underestimating higher canopy height values. 

Next, the predicted maps using our best performing model exhibit a similar pattern to the ALS reference maps; lower values correspond to water bodies and deforested areas, while higher values are observed in areas with dense forest cover. However, the lowest values are overestimated compared to those of ALS. These results align with the analysis conducted in the scatterplots, which indicated challenges in estimating heights lower than 4 m. In contrast, for forested regions, the model estimates are closer to the reference values. In fact, in most areas with high canopy height values in the ALS data, the model predictions are very close, indicating that the weighting strategy adopted to penalize fewer representative samples during the optimization process helped to improve the model’s capacity to predict values within that range. However, further research will be needed to address these challenges, especially when considering scenarios with higher canopy heights, such as tropical forests. In such cases, additional approaches may need to be evaluated to reduce the observed gaps in estimation accuracy. 

The final columns depict the associated calibrated uncertainty maps generated from the predictions. These maps elucidate regions characterized by higher confidence estimates alongside areas with higher uncertainty. Notably, the model demonstrates elevated confidence levels in regions adjacent to water bodies, with comparatively robust confidence levels observed in forested areas. Conversely, higher uncertainty is evident in border regions where transitions between forested areas and water bodies occur. This observation may be attributed to the presence of backscatter noise inherent to SAR sensors, a common occurrence in such scenarios. Furthermore, the accuracy of GEDI observations may also be compromised in these locations, potentially impacting the learning process of the neural network.

\section{Discussion}
The experimental findings of this study underscore the significant role of incorporating seasonal information from both optical and SAR data in enhancing canopy height estimates. Seasonal data is crucial for capturing phenological dynamics, which helps refine land cover characterization and improves model accuracy, particularly in environments with pronounced seasonal changes. By integrating these seasonal dynamics, models can more accurately represent variations in canopy height over time, leading to more precise and reliable height estimations. In this sense, consistent behavior was observed when comparing experiments employing seasonal and thus using exclusively summer data.  

Furthermore, the integration of multiple remote sensing data sources was found to substantially improve the model's capacity to estimate canopy height. Specifically, combining seasonal optical data with SAR data has demonstrated superior performance across key metrics, including R\textsuperscript{2}, RMSE, and MAE. The effectiveness of this approach is attributed to the complementary nature of the data types: optical data offers detailed spectral information, while SAR data, particularly from Sentinel-1, provides valuable structural insights (\cite{quan2024learning, sahour2021integrating}).  

The advantages of L-band SAR data, particularly its ability to penetrate forest canopies and extract valuable features related to tree stems, were evident in several experiments, especially those using only summer data. This capability greatly enhances the vertical resolution and the level of detail in forest structure assessments, thereby supporting more robust and accurate canopy height estimation. The integration of complementary radar frequencies, such as the L-band, indicates improvements in these models' reliability by providing a more comprehensive understanding of forest structures (\cite{koyama2019mapping, mitchell2014c, liesenberg2013evaluating}).  However, it is important to acknowledge certain limitations in the current study. The experiments relied on a one-year composite product due to the absence of consistent seasonal ALOS-PALSAR-2 data. This approach could lead to inconsistencies, as the variation in acquisition times across the study area might produce misleading results influenced by seasonal variability. Therefore, while the initial findings suggest that L-band SAR data contributes to canopy height estimation, a more comprehensive evaluation is necessary once consistent seasonal ALOS-PALSAR-2 data becomes available. This will help to address potential inaccuracies and provide a more precise understanding of L-band SAR data's role in different seasonal contexts. 

The contribution of different data inputs became particularly evident in experiments that relied exclusively on optical data collected during the summer. The findings suggest that this approach does not adequately capture the full range of canopy heights. Models trained using only summer data exhibited a higher bias compared to those that incorporated data from multiple seasons and other sensors. For instance, models using seasonal data showed significantly lower bias, whereas those limited to summer data alone demonstrated much higher bias under the most challenging conditions. This analysis underlines the critical importance of integrating data from various times of the year and complementary sources to enhance model performance, reduce bias, and provide a more accurate depiction of canopy height across different environments. 

Although the use of geographic coordinates is proven to be beneficial in global and national scales mapping (\cite{lang2023high, silveira2023nationwide, matasci2018mapping}), in this study the addition of such information had a detrimental effect on model performance for some cases. Latitude and longitude are in general used to capture trends associated with the regional temperature, precipitation and continentality gradients across the large scales, which may be useful to predicting canopy height, cover or biomass \cite{matasci2018mapping}. However, as our study was restricted to one province only, incorporating coordinates as predictor covariates may have compromised the models’ ability to generalize as it introduces spatial correlation dependency, making them more susceptible to local characteristics present in the training areas. 

When compared against the META product, our model exhibited consistently superior performance in both global metrics and visual map analyses. The META predictions tended to underestimate across the entire target data range, with a significant negative bias, as seen in the scatter plots. These results suggest potential domain-shift issues in the META models, likely due to a lack of supporting data from the evaluated region during their training. The lower coefficient of determination and higher error rates observed in the META product indicate that, while it can provide some useful insights, further refinement and the inclusion of additional region-specific data are needed to achieve the reliability demonstrated by GEDI’s estimates. 

Another important finding of this study indicates the effectiveness of using the negative log-likelihood of the Laplace distribution as a loss function for neural network optimization, which has been particularly effective at handling outliers \cite{nair2022maximum}. When comparing the scatter plots of GEDI observations with model predictions, it was noted that while GEDI maintained a stronger linear relationship with reference data, it produced more outliers than some models, resulting in a RMSE of approximately 4 meters. This discrepancy may result from acquisition errors due to weather conditions or geolocation errors, as documented in (\cite{lahssini2022influence, fayad2021quality, tang2023evaluating, weber2024unified}). In contrast, models trained with seasonal information consistently achieved RMSE values below 4 meters. These results suggest that, although GEDI provides a reliable baseline, trained models offer enhanced accuracy and consistency for height estimation tasks and facilitate spatially continuous canopy height mapping. While this level of precision might not be sufficient for applications requiring very high accuracy, it is generally adequate for a wide range of ecological and environmental studies. For example, in contexts like large-scale forest monitoring, carbon stock assessments, and biodiversity conservation, this error margin can still yield valuable insights and support effective decision-making. 

The study revealed significant challenges in accurately estimating canopy heights below 4 m, as indicated by the scatterplots comparing GEDI observations, model predictions, and reference data. These difficulties likely stem from intrinsic characteristics of the GEDI sensor that influence the trained models, limiting their ability to precisely predict lower height values. Such inaccuracies may be linked to errors in terrain height estimation \cite{adam2020accuracy}. Similar challenges were noted by \cite{travers2024mapping} in their study of northern forest areas in Canada using ICESat-2 data and random forest models, where canopy heights under 2 m were frequently overestimated. One potential explanation for these overestimations is the increased ground contribution to the signal in areas with low and sparse vegetation, which affects the accuracy of height measurements (\cite{schlund2019canopy, schlund2022towards}. Despite these challenges, under the global definitions of "Forest" “Land spanning more than 0.5 hectares with trees higher than 5 metres and a canopy cover of more than 10 percent, or trees able to reach these thresholds in situ” ~\cite{eggleston20062006, fao2020global}. Therefore, our focus was on enhancing model accuracy for taller canopies. In this context, our model demonstrated a higher saturation point compared to previous studies that reported underestimation of canopy heights around 25 to 30 m ( \cite{hansen2016mapping, healey2020highly, potapov2021mapping, sothe2022spatially}). As shown in Figure~\ref{fig:scatter_plots}, canopy heights exceeding 30 m are rare in our study area, and our models closely aligned with ALS data in predicting these values. This study, like \cite{lang2023high}, used a weighting strategy to penalize errors in less frequent height values, resulting in improved accuracy for higher canopy values. However, consistent with \cite{lang2023high}, our models still tended to overestimate shorter trees (< 5 m). This tendency is likely due to the limitations of GEDI in estimating low target values, which can sometimes lead to outliers ~\cite{dhargay2022performance}. 

Furthermore, the sparse sampling pattern of GEDI, as anticipated, limits its ability to comprehensively characterize the study area, leading to the omission of significant information, as reflected in the generated maps. This limitation highlights the need to integrate other satellite data to achieve continuous canopy height maps. Our analysis showed that the model estimates were generally closer to ALS data, albeit with a tendency to overestimate lower targets. The uncertainty maps also indicated lower confidence in areas at the boundaries, especially where there are transitions between forested areas and water bodies. These findings are consistent with those reported by~\cite{weber2024unified}, who described the effects of slopes on GEDI measurements. 

Most studies using LiDAR and spatially continuous covariates to estimate forest canopy height do not provide an uncertainty map (e.g., \cite{potapov2021mapping, sothe2022spatially, wagner2024sub}). Knowing the spatial distribution of uncertainty is crucial when important decisions are made based on the map content. For example, areas associated with higher levels of uncertainty can be disregarded from decision-making process, or this can prompt users or decision makers to collect in situ samples in high uncertainty areas. Here, we built uncertainty maps using an ensemble of neural networks trained to estimate a Laplace distribution, so that the output of each network are both the expected pixel value and uncertainty value. By adopting this approach, like \cite{lang2023high} we are able to capture the epistemic and aleatoric uncertainty. 

\section{Conclusions}

Our study demonstrates the effectiveness of leveraging multiseasonal data and advanced machine learning techniques to enhance canopy height estimation using GEDI observations in Ontario, Canada. By employing a Fully Convolutional Neural Network (FCNN) and incorporating uncertainty estimates through the minimization of the negative log-likelihood of the Laplacian, we achieved notable improvements in accuracy and robustness. The results underscore the significant benefits of integrating seasonal information from both optical and SAR data. Using seasonal data instead of summer-only data increased explained variability by 10\%, reduced canopy height error by 0.45 m, and decreased bias by 1 m. Seasonal data helps capture phenological dynamics, thereby enhancing land cover characterization and model accuracy, particularly in regions with pronounced seasonal variations. The combined use of seasonal optical and SAR data demonstrated superior performance across the evaluated metrics due to the complementary strengths of these data types—optical data providing rich spectral details and SAR data delivering valuable structural insights. 
Conversely, the worst performance was observed when only optical and summer data were used, highlighting the necessity of using complementary data sources to fully characterize the problem. This configuration resulted in a 22\% reduction in the model's capacity to capture data variability, reinforcing the importance of integrating diverse data inputs. Future research could focus on incorporating additional datasets, such as seasonal ALOS-PALSAR-2 and ICESat-2 data, to further refine canopy height estimates. Exploring data fusion strategies that combine GEDI and ICESat-2 observations may also enhance model performance. Additionally, expanding the study area to include more diverse forest types across Canada could provide a more comprehensive assessment. Overall, our study advances canopy height estimation methodologies and offers valuable insights for remote sensing applications in forestry and environmental monitoring.

\section*{Acknowledgement}
This research was funded by the Natural Sciences and Engineering Research Council of Canada (NSERC), Canada Research Chairs (CRC2019-00139) program and Environment and Climate Change Canada.

The findings and views described herein do not necessarily reflect those of Planet Labs PBC.

\printcredits

\bibliographystyle{cas-model2-names}

\bibliography{cas-refs}






\begin{algorithm*}[t]
\caption{Seasonal Image Composite}\label{alg:composite}

\begin{algorithmic}
\Require $Y$: Year; $SN$: Season; $SR$: Sensor; $ROI$: Region of Interest; $N$: Minimum number of observation at each pixel location; $W$: Number of years back considered.
\\
\State 1. $I_{targetCollection}^{(0)} \gets \emptyset_{m \times n} $
\Comment{initialize the target collection with an empty image collection, where $m$ and $n$ represent the vertical and horizontal $ROI$ spatial dimension, respectively.}

\While{$\omega \leq W$}:

\State 2. $I_{counter} \gets pixel\_counter(I_{targetCollection})$ \Comment{count the number of valid observations available at each pixel location. } 

\State 3. $I_{Collection} \gets get\_image\_collection(Y, SN, SR, ROI)$ \Comment{gather all the images matching with $Y$, $SN$, $SR$, and $ROI$ queries. }

\If{$SR$ is optical}:
\State $I_{Collection} \gets masking\_pixels(I_{Collection}, SN)$\Comment{mask out noise pixels, i.e., clouds, shadows, snow (depending on the seasons), etc.}
\EndIf

\State 4. $I_{targetCollection}^{(w)}  \gets concat\_collections([I_{targetCollection}, I_{Collection}], I_{counter})$\Comment{ concatenate image collections only at pixel locations with less than $N$ valid observations.}

\State 5. $Y \gets Y - 1 $ \Comment{update the target year}

\EndWhile
\State 6. $I_{comp} \gets med(I_{targetCollection}^{(w)})$ \Comment{ compute the median image.}
\\
\Return $I_{comp}$

\end{algorithmic}
\end{algorithm*}

\end{document}